\documentclass[letterpaper, 10 pt, journal, twoside]{IEEEtran}
\ifCLASSINFOpdf
\else
\fi
\usepackage{url}
\usepackage{overpic}
\usepackage{xcolor}
\usepackage{amsmath,amssymb,amsfonts}
\usepackage{algorithm}
\usepackage[noend]{algpseudocode}
\usepackage{colortbl}
\usepackage[margin=0.79in, left=0.67in, right=0.67in, bottom=0.6in]{geometry}
\definecolor{myBlue}{HTML}{0000FF}
\definecolor{myPink}{HTML}{FF69B4}
\definecolor{myCyan}{HTML}{00B3B3}

\begin{document}
\title{A Genetic Approach to Gradient-Free Kinodynamic Planning in Uneven Terrains}

% \author{Michael~Shell,~\IEEEmembership{Member,~IEEE,}
%         John~Doe,~\IEEEmembership{Fellow,~OSA,}
%         and~Jane~Doe,~\IEEEmembership{Life~Fellow,~IEEE}% <-this % stops a space
% \thanks{M. Shell was with the Department
% of Electrical and Computer Engineering, Georgia Institute of Technology, Atlanta,
% GA, 30332 USA e-mail: (see http://www.michaelshell.org/contact.html).}% <-this % stops a space
% \thanks{J. Doe and J. Doe are with Anonymous University.}% <-this % stops a space
% \thanks{Manuscript received April 19, 2005; revised August 26, 2015.}}
\author{Otobong Jerome$^{1}$, Alexandr Klimchik$^{2}$, Alexander Maloletov$^{1}$, and Geesara Kulathunga$^{2}$%
\thanks{Manuscript received: December 24, 2024; Revised February 9, 2025; Accepted March 25, 2025.} %Use only for final RAL version
\thanks{This paper was recommended for publication by Editor Aniket Bera upon evaluation of the Associate Editor and Reviewers' comments.
} %Use only for final RAL version
\thanks{$^{1}$Otobong Jerome and Alexander Maloletov are with the Center of Autonomous Technologies, Innopolis University, Innopolis, Russia.
{\tt\small \{o.jerome, a.maloletov\}@innopolis.university}}%
\thanks{$^{2}$Alexandr Klimchik and Geesara Kulathunga are with the School of Computer Science, University of Lincoln, United Kingdom.
{\tt\small \{gkulathunga, aklimchik\}@lincoln.ac.uk}}%
\thanks{Digital Object Identifier (DOI): see top of this page.}
}

\markboth{IEEE Robotics and Automation Letters. Preprint Version. Accepted March, 2025}
{Jerome \MakeLowercase{\textit{et al.}}: Genetic Kinodynamic Planning}




\maketitle

\begin{abstract}
This paper proposes a genetic algorithm-based kinodynamic planning algorithm (GAKD) for car-like vehicles navigating uneven terrains modeled as triangular meshes. The algorithm's distinct feature is trajectory optimization over a receding horizon of fixed length using a genetic algorithm with heuristic-based mutation, ensuring the vehicle’s controls remain within its valid operational range. By addressing the unique challenges posed by uneven terrain meshes, such as changes face normals along the path, GAKD offers a practical solution for path planning in complex environments. Comparative evaluations against the Model Predictive Path Integral (MPPI) and log-MPPI methods show that GAKD achieves up to a 20\% improvement in traversability cost while maintaining comparable path length. These results demonstrate the potential of GAKD in improving vehicle navigation on challenging terrains.
\end{abstract}

% \begin{IEEEkeywords}
% IEEE, IEEEtran, journal, \LaTeX, paper, template.
% \end{IEEEkeywords}

% \fi
\IEEEpeerreviewmaketitle

\section{Introduction}
% for IEEE journal papers produced under \LaTeX\ using
% IEEEtran.cls version 1.8b and later.
\IEEEPARstart{K}{inodynamic} motion planning is a critical area in robotics that involves generating collision-free trajectories while respecting a robot’s kinematic and dynamic constraints. Kinodynamic planning incorporates constraints such as acceleration and force, significantly increasing the complexity of the problem. Accounting for dynamic constraints requires integrating knowledge of the control system and optimization into the pathfinding problem, thereby linking robotics, dynamical systems, and control theory.

Kinodynamic motion planning was first formalized in \cite{L1} and \cite{L2}. Other early work introduced foundational algorithms, such as Rapidly-Exploring Random Trees (RRTs) \cite{L4}, which became a standard in the field. The field was further advanced by RRT variants, such as RRT*, which incorporates cost optimization alongside dynamic feasibility \cite{L5}.

Beyond RRTs, modern gradient-free approaches such as Model Predictive Path Integral (MPPI) \cite{williams2016aggressive}, represent significant advancements in motion planning by handling the complexities of nonlinear dynamics and real-time computation. 
Although Genetic Algorithms (GAs) have been widely used in path planning, their application in an uneven terrain setting with kinodynamic constraints has not caught enough attention of the research community.
Hence, the main goals of this paper are:
\begin{itemize}
    \item A mathematical model of a car-like vehicle motion on a triangular mesh environment.
    \item An efficient genetic algorithm-based, and gradient-free algorithm for computing
kinodynamically efficient path to a goal from an arbitrary point on a 3D triangular mesh surface.
    \item A Robot Operating System (ROS) planner package that integrates with Move Base Flex (MBF) \cite{L36} and implements the proposed algorithm.
\end{itemize}
\begin{figure}[htbp]
    \centering
    \begin{overpic}[width=0.8\linewidth]{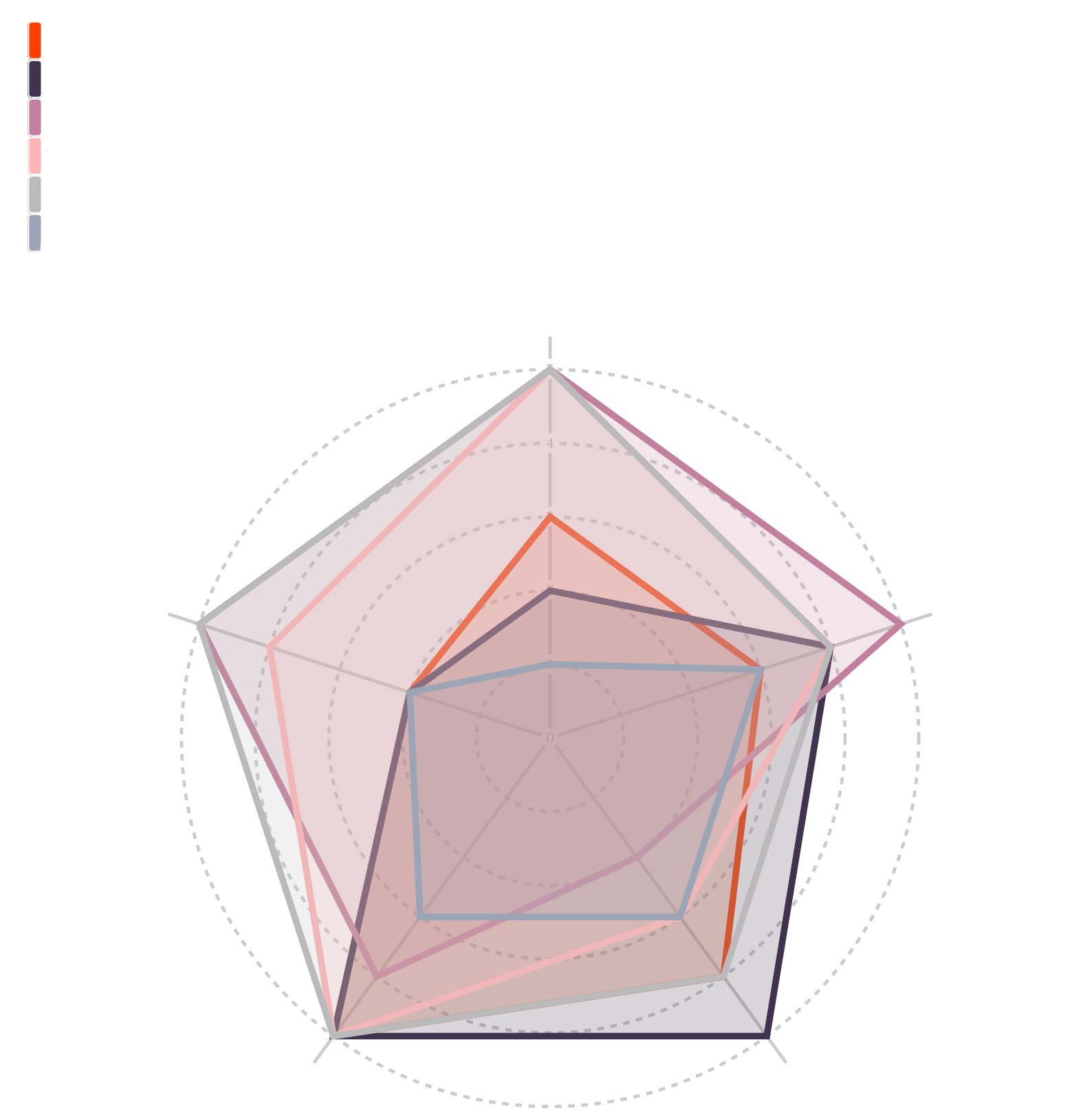}
        \put(5, 95.5){Rapidly-exploring Random Trees (RRTs)} 
        \put(5, 91.5){Iterative Linear Quadratic Regulator (ILQR)} 
        \put(5, 88){Covariant Hamiltonian Optimization (CHOMP)} 
        \put(5, 84.5){Model Predictive Control (MPC)} 
        \put(5, 81){Model Predictive Path Integral Control (MPPI)} 
        \put(5, 77){Probabilistic Roadmaps}
        \put(20, 70){\color{myBlue} Nonlinear Dynamics And Constraints} 
        \put(77, 47){\color{myBlue} Optimality} 
        \put(1, 56){\color{myBlue} Environment}
        \put(1, 52){\color{myBlue} Complexity} 
        \put(-3, 0){\color{myBlue} Real-Time Applicability} 
        \put(47, 0){\color{myBlue} Initial Conditions Sensitivity} 

    \end{overpic}
    \caption{A radar chart comparing the performance of various motion planning methods across key criteria. While sampling-based approaches may lack global optimality, they are fast and practical, excelling in complex environments.}
    \label{fig:enter-label}
\end{figure}


\section{Related work}
\begin{figure*}
  \centering
  \begin{overpic}[width=0.8\linewidth]{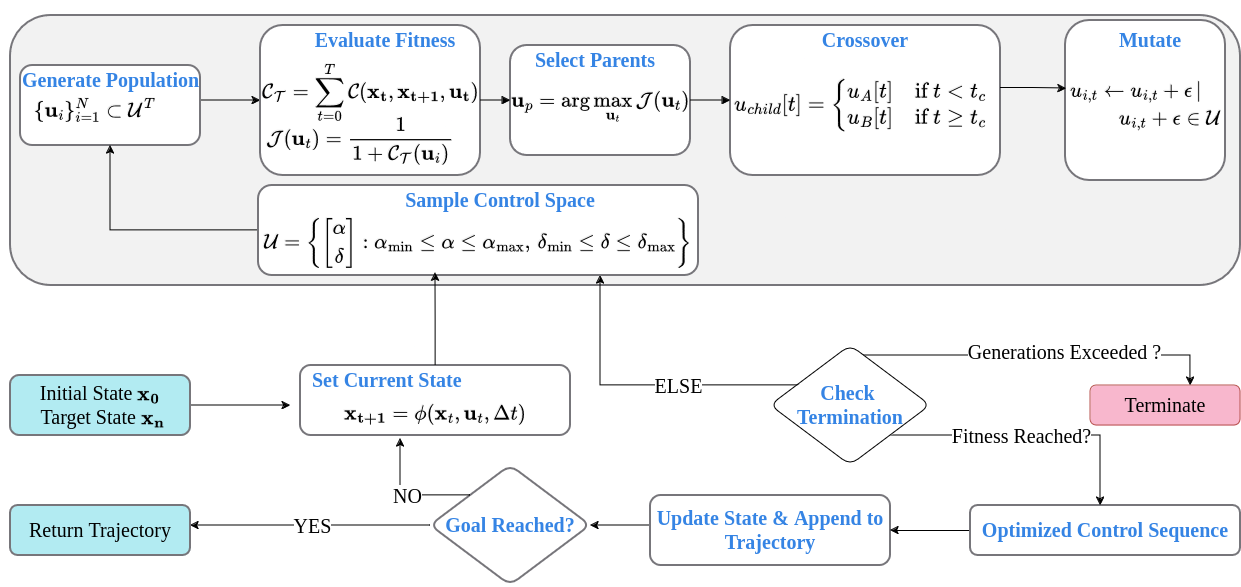}
  \end{overpic}
  \caption{Flow chart of the proposed genetic algorithm-based kinodynamic planning: GA is used to find the sequence of controls with the best fitness while ensuring all the controls in the sequence are from the space of valid controls.}
  \label{fig:abstract}
\end{figure*}
Kinodynamic path planning addresses the challenge of computing collision-free trajectories for mobile robots or agents, considering both kinematic and dynamic constraints. The earliest formalisms of the problem include \cite{L1} and \cite{L2}. In \cite{L3} a polynomial-time $\epsilon$-approximation algorithm is proposed. These early methodologies set the foundation for kinodynamic motion planning, and subsequent innovations focused on solving this complex problem more efficiently.
One common method is Rapidly-exploring Random Trees (RRTs) introduced by \cite{L4}, a variant called RRT* as well as probabilistic roadmaps proposed by \cite{L5}. These algorithms optimize the path toward a minimum-cost solution, addressing both feasibility and optimality in kinodynamic spaces.

Gradient-based approaches such as Sequential Quadratic Programming (SQP) \cite{torrisi2016variant}, \cite{bomze2010sequential}, \cite{ansary2020sequential},  Model Predictive Control (MPC) \cite{pari2013model}, \cite{lam2012model}, \cite{ren2010path}, Covariant Hamiltonian Optimization for Motion Planning (CHOMP) \cite{zucker2013chomp}, \cite{ratliff2009chomp}, and Iterative Linear Quadratic Regulator (ILQR) \cite{van2014iterated} are widely employed to ensure that trajectories respect dynamic constraints. These methodologies share a common objective: however, they differ significantly in their operational mechanisms, associated advantages, and inherent limitations.

Sequential Quadratic Programming (SQP) and Iterative Linear Quadratic Regulator (ILQR) are both iterative methods aimed at solving optimization problems, but they approach the task differently. SQP solves nonlinear problems by breaking them into a series of quadratic programming problems \cite{torrisi2016variant}, \cite{bomze2010sequential}. SQP is known for its fast convergence but is highly sensitive to the initial conditions, and may only find local optima, making it less ideal for problems requiring global solutions\cite{bomze2010sequential}, \cite{ansary2020sequential}. In contrast, ILQR focuses on improving trajectories by linearizing system dynamics and approximating the cost function as quadratic \cite{van2014iterated}. While ILQR is computationally efficient, especially for large-scale systems, it may struggle when handling highly nonlinear or non-convex constraints, thus limiting its effectiveness in certain complex environments.

Comparatively, CHOMP and MPC offer distinct advantages in terms of their capacity to handle dynamic systems, but with different approaches. The difference is that CHOMP combines Hamiltonian mechanics with optimization techniques to generate motion plans that are both efficient and physically compliant \cite{zucker2013chomp}, \cite{ratliff2009chomp}. CHOMP excels in high-dimensional spaces and produces smooth, stable trajectories, making it particularly useful in robotics and autonomous systems. Unlike SQP and ILQR, which rely on linear or quadratic approximations, CHOMP’s focus on leveraging physical principles results in more globally robust solutions, though it may require higher computational resources in complex environments \cite{zucker2013chomp}, \cite{ratliff2009chomp}.

On the other hand, MPC takes a unique approach by optimizing over a finite horizon in real-time, adjusting at each time step based on current and predicted future states\cite{pari2013model}, \cite{lam2012model}. This makes MPC highly adaptive to dynamic environments and capable of handling time-varying constraints. However, its real-time applicability comes at a cost—MPC requires fast online computation, which can be computationally demanding, particularly in highly nonlinear scenarios\cite{pari2013model}, \cite{ren2010path}.

Model Predictive Path Integral Control (MPPI) is a sampling-based MPC algorithm that handles nonlinear dynamics and non-convex constraints effectively.
Williams et al. \cite{williams2016aggressive} demonstrated that MPPI enables real-time trajectory generation in robotic systems, balancing computational efficiency with high-quality trajectory outcomes. Unlike RRTs, which incrementally explore the state space, MPPI evaluates multiple control inputs in parallel since its generated trajectories are independent, making it better suited for systems with continuous dynamics, as demonstrated by \cite{9164598} and \cite{10336978}.
The effectiveness of the MPPI control method relies on choosing the right input distribution for sampling \cite{10611180}, leading to the creation of variants such as the Gaussian MPPI \cite{williams2016aggressive}, the Output-Sampled Model Predictive Path Integral Control (o-MPPI) \cite{10611180} and Biased-MPPI \cite{10520879}. 
There is a need for an approach that reduces reliance on the sampling distribution, focusing instead on a heuristic based on the known space of valid vehicle controls.
The proposed genetic algorithm uses a heuristic mutation strategy which allows an efficient exploration within the space of valid controls. GAs are search heuristics inspired by the process of natural selection, often used to find global optima in large and complex search spaces. While GAs have been employed for general robotic path planning, they primarily focus on geometric constraints without incorporating dynamic aspects such as velocity or force. For instance, \cite{tu2003genetic} and \cite{hu2004knowledge} explored the use of GAs in motion planning in a 2D plane setting. However, their use in kinodynamic settings, where dynamic feasibility is equally crucial, remains largely unexplored.
The methodology integrates GA with the existing framework of sampling based gradient free planning, and  demonstrates its applicability to car-like vehicle moving on an uneven terrain modeled by triangular meshes.

\section{Methodology}
The proposed algorithm aims to find a kinodynamically feasible path as a sequence of points $\mathbf{p} \in \mathbb{R}^3$ on the mesh surface. The two key constraints considered are as follows:
\begin{itemize}
    \item For car-like vehicles, the kinodynamic constraints limit the steering angle and acceleration.
    \item The change in orientation, quantified by the traversability cost, must remain within an allowable range.
\end{itemize}
A path is deemed feasible if, for each consecutive point-to-point transition from $\mathbf{p_i}$ to $\mathbf{p_{i+1}}$, the traversability cost is acceptable, a valid control exists to move the vehicle between these points, and the entire path remains on the mesh surface.

As shown in Figure \ref{fig:abstract}, the planning starts by initializing the current state from the initial state \(\mathbf{x_{\text{init}}}\). A population of potential control sequences is generated by sampling from the space of valid controls. Each sequence is evaluated based on its fitness, which reflects the cost of applying that control. The algorithm then selects the best-fit solutions, performs crossover and mutation to generate new offspring, and checks if the termination criterion (fitness threshold) is met. If the fitness is satisfactory, the optimized control sequence is returned. Otherwise, the process repeats with a new population. After each iteration, the current state is updated by simulating the system dynamics using the first control input from the optimized control sequence, and the resulting state is appended to the trajectory. The planning continues until the goal state is reached, at which point the trajectory is returned.

Below we describe the system dynamic model, used to estimate the behavior of a car-like vehicle moving on a 3D mesh. The proposed algorithm does not depend only on the specific dynamic model described here, various linear and nonlinear ones can be used.

\subsection{Dynamic Model of Car-like Vehicle on a Mesh}
For context, mesh generation falls outside the scope of the proposed methodology. Specifically, \cite{L36}, \cite{putz_mesh}, and \cite{lvr2}, were leveraged because these tools generate Navigation Meshes from point cloud scans. Navigation Meshes represent navigable areas in 3D environments, incorporating features, such as local roughness estimation, face normals, and obstacle marking, enabling direct planning.

Let  
\begin{equation}
\mathbf{x} = 
\begin{bmatrix}
    \mathbf{p} & \boldsymbol{\theta} & \mathbf{v}
\end{bmatrix}^\intercal, \quad
\mathbf{u} = 
\begin{bmatrix}
    a & \delta
\end{bmatrix}^\intercal
\end{equation}
$\mathbf{x}$ represent the vehicle's state, where:
\[
\mathbf{p} = 
\underbrace{\begin{bmatrix} x & y & z \end{bmatrix}}_{\text{Position}} 
\boldsymbol{\theta} = \underbrace{\begin{bmatrix} \theta_{\text{yaw}} & \theta_{\text{pitch}} & \theta_{\text{roll}} \end{bmatrix}}_{\text{Orientation}}
\mathbf{v} = \underbrace{\begin{bmatrix} v_x & v_y & v_z \end{bmatrix}}_{\text{Velocity}}
\]
$\mathbf{u}$ represent the control input, where \(a \in \mathbb{R}\) is the linear acceleration and \(\delta \in \mathbb{R}\) is the steering angle (yaw control).
We describe the change of the state \(\mathbf{x_t} = [\mathbf{p}, \boldsymbol{\theta}, \mathbf{v}]^\intercal\) when the control input \(\mathbf{u_t} = [a, \delta]^\intercal\) is applied for a time step size \(\Delta t\).
\subsection*{Assumptions}
\begin{itemize}
    \item  The time interval \(\Delta t\) is chosen such that the vehicle moves to at most one adjacent face at a time, we used 0.1 seconds for the experiments.
   \item The normal vectors \((n_x, n_y, n_z)\) of the mesh triangular faces are known, derived from LiDAR scans of the terrain.
\item The positions of the centers \(\mathbf{p}_{\text{center}}\) of the mesh's triangular faces are known from the LiDAR-generated mesh data.

    \item Gravity acts along the z-axis of the global frame.
    \item Friction depends on the vehicle’s velocity along the mesh surface.
    \item The mass of the vehicle is assumed to be 1 (unit mass for simplicity).
\end{itemize}
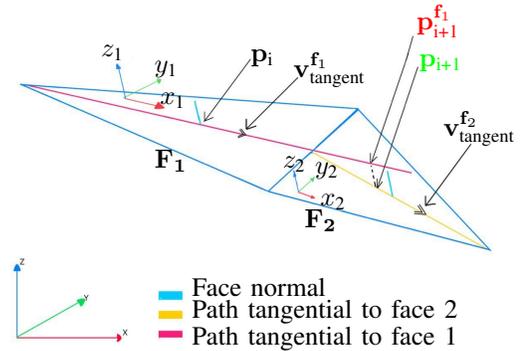
\begin{figure}[ht]
    \centering
    \begin{overpic}[width=0.8\linewidth]{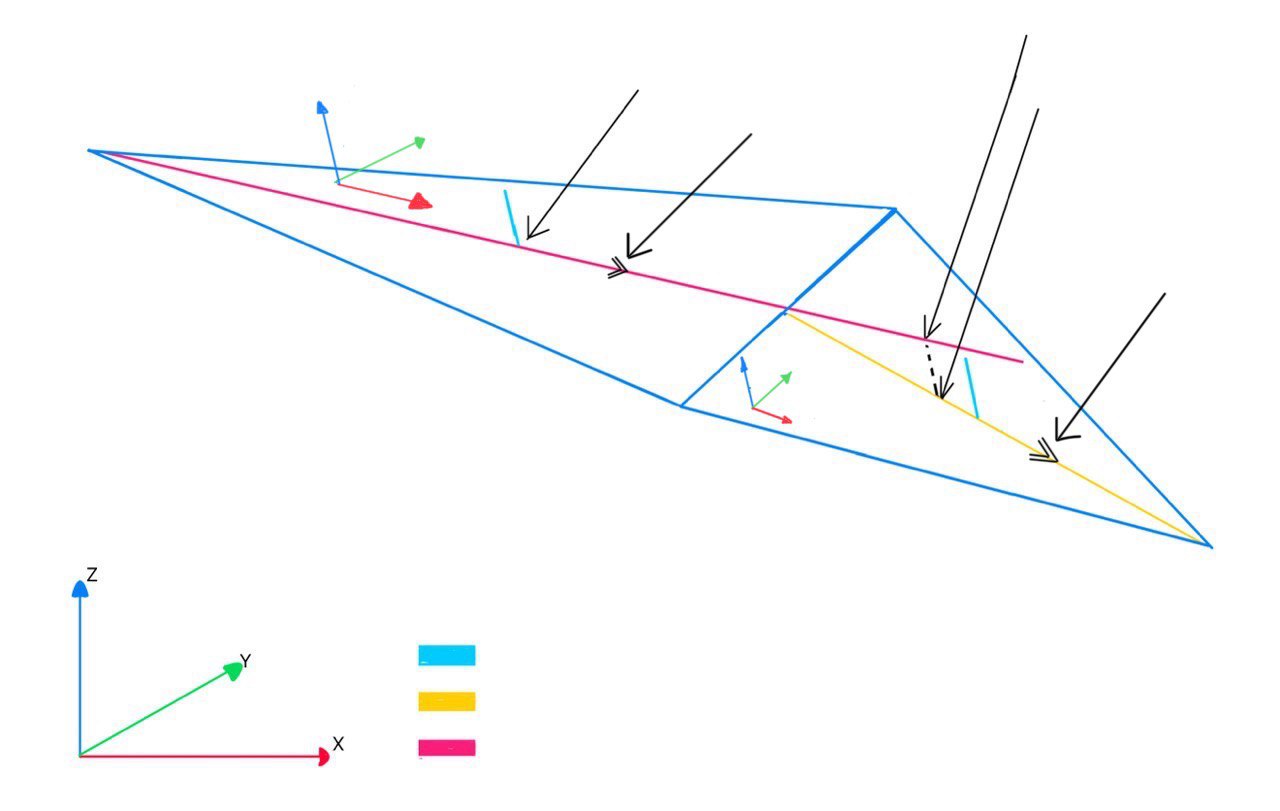}
        \put(60, 25){$\mathbf{F_2}$} 
        \put(32, 36){$\mathbf{F_1}$} 
        \put(39, 12){Face normal} 
        \put(39, 8){Path tangential to face 2} 
        \put(39, 3){Path tangential to face 1}
        \put(22, 57){${z}_{\text{1}} $}
        \put(32, 54){${y}_{\text{1}} $}
        \put(33, 48){${x}_{\text{1}} $}
        \put(56, 36){${z}_{\text{2}} $}
        \put(62, 35){${y}_{\text{2}} $}
        \put(63, 29){${x}_{\text{2}} $}
        \put(50, 56){$\mathbf{p}_{\text{i}} $}
        \put(58, 53){$\mathbf{v}_{\text{tangent}}^{\mathbf{f}_1} $}
        \put(81, 62){\color{red} $\mathbf{p}_{\text{i+1}}^{\mathbf{f}_1} $} 
        \put(81.5, 55){\color{green} $\mathbf{p}_{\text{i+1}} $} 
        \put(86, 42){$\mathbf{v}_{\text{tangent}}^{\mathbf{f}_2} $} 
    \end{overpic}
    \caption{If a vehicle moves from \(\mathbf{p_i}\) tangentially along face \(\mathbf{F_1}\) to \(\mathbf{p}_{i+1}^{\mathbf{f_1}}\), it has a tangential velocity \(\mathbf{v}_{\text{tangent}}^{\mathbf{f}_1} \). Upon crossing into face \(\mathbf{F_2}\), the next position is projected to \(\mathbf{p}_{i+1}\) with a new tangential velocity \(\mathbf{v}_{\text{tangent}}^{\mathbf{f}_2} \). The vehicle reference frame undergoes an orientation change from face \(\mathbf{F_1}\) to face \(\mathbf{F_2}\), with coordinates \(({x}_1, {y}_1, {z}_1)\) on \(\mathbf{F_1}\) transforming to \(({x}_2, {y}_2, {z}_2)\) on \(\mathbf{F_2}\).}

    \label{fig:projection}
\end{figure}
\paragraph{Change In Position}
The vehicle's position is influenced by its tangential velocity, given by 
\begin{equation}
    \label{vtangent}
    \mathbf{v}_{\text{tangent}} = \mathbf{v} - (\mathbf{v} \cdot \mathbf{\hat{n}}) \mathbf{\hat{n}}
\end{equation},
where \( \mathbf{\hat{n}} \) is the normal of the current face. The forward direction is the unit vector along the tangential velocity, leading to the updated position after a time step \( \Delta t \):
\begin{equation}
    \mathbf{p}_{\text{i+1}}^{\mathbf{f_i}} = \mathbf{p}_{\text{i}} + \mathbf{v}_{\text{tangent}} \Delta t.
\end{equation}
If the updated position crosses into a neighboring face \( \mathbf{F_2} \), it is projected onto the face to ensure movement remains constrained to the mesh surface. The tangential velocity on the new face can be estimated similarly using \eqref{vtangent} and the normal of the new face.
\paragraph{Projection onto Mesh $\mathcal{M}$}
Given a face \( \mathbf{F_2} \) with normal vector \( \mathbf{n} = (a, b, c) \) and center \( \mathbf{p}_{\text{center}} \), the face equation is \( ax + by + cz + d = 0 \), where \( d = -\mathbf{n} \cdot \mathbf{p}_{\text{center}} \). The signed distance from the point \( \mathbf{p}_{\text{i+1}}^{\mathbf{f_i}} \) to the plane is:
\begin{equation}
    \label{eqn:gamma}
    \gamma = \frac{ax_0 + by_0 + cz_0 + d}{\sqrt{a^2 + b^2 + c^2}}.
\end{equation}
The projected point is then computed as:
\begin{equation}
    \mathbf{p}_{\text{i+1}} = \mathbf{p}_{\text{i+1}}^{\mathbf{f_i}} - \gamma \mathbf{n}.
\end{equation}
This ensures that the vehicle remains on the mesh surface throughout its motion.
\paragraph{Change in Velocity}
The vehicle's velocity is influenced by applied acceleration, friction, and gravity. The control input \( a \) is applied along the vehicle's forward direction \(\cdot \mathbf{\hat{d}_{\text{forward}}} = \frac{\mathbf{v}_{\text{tangent}}}{\|\mathbf{v}_{\text{tangent}}\|} \), leading to a tangential acceleration: \(\mathbf{a_{\text{tangent}}} = a \cdot \mathbf{\hat{d}_{\text{forward}}}.\) Friction decelerates the vehicle along its velocity direction, given by: \(\mathbf{a_{\text{friction}}} = \mu g \cdot \mathbf{\hat{d}_{\text{forward}}},\)
where \( \mu \) is the friction coefficient and \( g \) is gravitational acceleration. Gravity influences the velocity along the surface normal: \(\mathbf{a_{\text{gravity}}} = -g \cdot \mathbf{n}.\)
The updated velocity is then:
\begin{equation}
    \mathbf{v}_{\text{next}} = \mathbf{v} + (\mathbf{a_{\text{tangent}}} - \mathbf{a_{\text{friction}}} + \mathbf{a_{\text{gravity}}}) \Delta t.
\end{equation}

\paragraph{Change in Orientation}
The vehicle’s orientation, described by yaw, pitch, and roll, is updated based on steering and surface normal. The yaw rate due to steering angle \( \delta \) is: \(\omega_z = \frac{v}{L} \tan(\delta),\)
where \( L \) is the wheelbase. The yaw angle is updated as:
$\theta_{\text{yaw, next}} = \theta_{\text{yaw}} + \omega_z \Delta t.$
Pitch and roll angles correspond to the surface's slope and tilt:
$\theta_{\text{pitch}}, \theta_{\text{roll}} = \text{atan2} ( n_z, \sqrt{n_x^2 + n_y^2} ), \text{atan2} ( n_y, n_z ).$
Putting it together, the state transition can be described as follows:
\begin{equation}
\label{eqn:dynamics}
\mathbf{ x_{t+1}} = \phi(\mathbf{x}_t, \mathbf{u}_t, \Delta t)   
\end{equation}

where:
\begin{equation}
    \phi(\mathbf{x}_t, \mathbf{u}_t, \Delta t) = \begin{bmatrix}
\mathcal{M}(\mathbf{p}_{t} + \mathbf{v}_{\text{tangent}} \Delta t) \\
\theta_{\text{yaw}, t} + \frac{v_t}{L} \tan(\delta_t) \Delta t \\
\mathbf{v} + (\mathbf{a_{\text{tangent}}} - \mathbf{a_{\text{friction}}} + \mathbf{a_{\text{gravity}}}) \Delta t
\end{bmatrix}
\end{equation}.

Given $\mathbf{x}_t$, $\mathbf{u}_t$, and $\Delta t$, the dynamics estimate the next state $\mathbf{x}_{t+1}$ with an associated error, which can be estimated using \eqref{eqn:gamma}. This error can be reduced by selecting a smaller $\Delta t$ and adjusting the control input $\mathbf{u}_t$.
Additionally, the vehicle's orientation is determined by the normal of the surface it is on. For two adjacent surfaces along the path, it is preferable for the normals to be relatively aligned. To enforce this, we introduce a penalty $\Pi$ \eqref{eqn:travcost}.

\begin{figure}[H]
    \centering
    \begin{overpic}[width=0.7\linewidth]{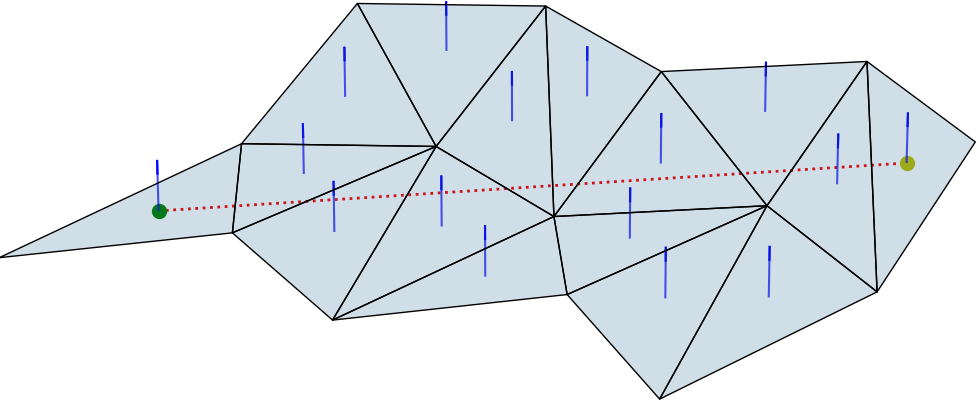}
        \put(13, 20){$\mathbf{P}$} 
        \put(87, 23){$\mathbf{P_{target}}$} 
    \end{overpic}
    \label{fig:cost}
    \caption{The blue arrows represent the face normals, while the dotted line represents the direction vector \(\mathbf{j}\) from \(\mathbf{P}\) to \(\mathbf{P_{\text{target}}}\). In an ideal scenario where all faces are flat, the face normals along the path should be orthogonal to \(\mathbf{j}\), the deviation from this which is penalized by \(\Sigma\). For adjacent faces, if they are perfectly aligned, their normals should be parallel, and deviations from this are penalized by \(\lambda\). The combination of \(\Sigma\) and \(\lambda\) defines \(\Pi\), the traversability cost.}
\end{figure}

\subsection{Preprocessing and Voxelization}
In sequential planning algorithms, projecting each step onto a mesh increases time complexity. To optimize this, methods like barycentric coordinates and K-D trees are commonly used. Barycentric coordinates identify the enclosing face, while K-D trees enable fast nearest-neighbor searches.
For constant-time face normal retrieval, the mesh is voxelized, with each voxel storing the nearest face center and normal in a hash table. Voxel coordinates $(i, j, k)$ are computed as follows:
$i, j, k = \left\lfloor \frac{x - x_{\text{min}}}{\text{voxel\_size}} \right\rfloor, \left\lfloor \frac{y - y_{\text{min}}}{\text{voxel\_size}} \right\rfloor, \left\lfloor \frac{z - z_{\text{min}}}{\text{voxel\_size}} \right\rfloor $.
The hash map is then queried for the precomputed normal: $\mathbf{n} = \text{hash\_map.get}(i, j, k)$. This preprocessing step enables efficient normal retrieval, reducing computational overhead.
\subsection{Genetic Algorithm based Kinodynamic Planning}
The optimization strategy employed a GA to find an optimal control sequence over a finite horizon, aimed at minimizing a given cost function while evolving the system dynamics.

\begin{itemize}
\item \textbf{Initial Population:} We initialized a valid control space \(\mathcal{U}\) from which we sampled control sequences, each representing a potential policy over the time horizon. Control values were constrained by the vehicle's limits:
$\mathcal{U}  = \left\{
\begin{bmatrix}
\alpha \\ \delta
\end{bmatrix} : \alpha_{\text{min}} \leq \alpha \leq \alpha_{\text{max}}, \, \delta_{\text{min}} \leq \delta \leq \delta_{\text{max}}
\right\}$.
We sample multiple control sequences, ensuring each control \( u \) in the sequence is drawn from \(\mathcal{U}\).
\item \textbf{State Transition and Dynamics:} For each control sequence in the population, the system was rolled out starting from an initial state \( \mathbf{x_{\text{init}}} \). The system dynamics \eqref{eqn:dynamics} was computed for each time step using the control input from the sequence:
 \item \textbf{Cost Function:} The cost function evaluates each state transition \( (\mathbf{x_t} \to \mathbf{x_{t+1}}) \) based on the state and control input.
The cost function \( \mathcal{C} \) for a state transition was defined as follows:
    \begin{equation}
    \label{eqn:cost}
            \mathcal{C} = \alpha_1\Delta + \alpha_2\Pi 
    \end{equation}
    The total cost over the finite time horizon $T$ was computed as follows:
    \begin{equation}
        \label{eqn:total_cost}
        \mathcal{C_T} = \sum_{t=0}^T \mathcal{C}(\mathbf{x_t}, \mathbf{x_{t+1}}, \mathbf{u_t})
    \end{equation}
    In this context:
    \(\alpha_1\),\(\alpha_2\) are weight parameters
$\Delta = \| \mathbf{p}_{\text{i+1}} - \mathbf{p}_{\text{target}} \|$
In this equation, \(\mathbf{p}\) represents the position vector components of the state \(\mathbf{x}\). The term \(\Delta\) represents the distance cost, the distance between the next state position \(\mathbf{p}_{\text{i+1}}\) and the target position \(\mathbf{p}_{\text{target}}\).
    \begin{equation}
    \label{eqn:travcost}
            \Pi = \frac{1}{2} \left( \Sigma + \Lambda \right)
    \end{equation}
    \(\Pi\) represents the traversability cost, penalizing the difficulty of the transition between adjacent state positions in terms of the terrain's topology. It consists of: $\Sigma = |\mathbf{n}_{\text{t}} \cdot \mathbf{j}|,  \quad   \text{and} \quad \Lambda = 1 - |\mathbf{n}_{\text{t}} \cdot \mathbf{n}_{\text{t+1}}|$
    \(\Sigma\) is the slope penalty, which measures the alignment of the path vector \(\mathbf{j} = \mathbf{p}_{\text{t+1}} - \mathbf{p}_{\text{target}}\) with the normal vector \(\mathbf{n}_{\text{t}}\) at the current mesh face. Higher values indicate a steeper slope and result in higher penalties.
     \(\Lambda\) is the orthogonality penalty that discourages significant changes in terrain. It does this by measuring the difference in terrain normals between the current state \(\mathbf{n}_{\text{t}}\) and the next state \(\mathbf{n}_{\text{t+1}}\). The value of \(\Lambda\) approaches 0 when the normals are aligned and approaches 1 when the normals are orthogonal.
    \item \textbf{Fitness Assignment:}
    The fitness $\mathcal{J}(\mathbf{u}_t)$ of each control sequence $\mathbf{u}_t$ is determined by its total cost. $\mathcal{J}(\mathbf{u}_t) = \frac{1}{1 + \mathcal{C_T}(\mathbf{u}_t)}$
    Hence, a lower cost indicates better fitness. 
   \item \textbf{Parent Selection and Crossover:} Two parent control sequences are selected using a tournament selection method based on fitness, balancing performance and diversity. Crossover is then performed to produce two offspring by combining segments of these parent sequences, aiming for enhanced solutions. The child sequence \( u_{c} \) is generated from parent sequences \( u_A \) and \( u_B \) with a crossover threshold \( t_c \) as follows: $ u_{c}[t] =
    \begin{cases}
    u_A[t] & \text{if } t < t_{c} \\
    u_B[t] & \text{if } t \geq t_{c}
    \end{cases}$,
    This mechanism sets \( u_{c} \) to take values from \( u_A \) until \( t = t_c \), then from \( u_B \) starting at \( t = t_c \).
    
\begin{figure}[htbp!]
    \centering
    \vspace{0.5cm}
    \begin{overpic}[width=0.7\linewidth]{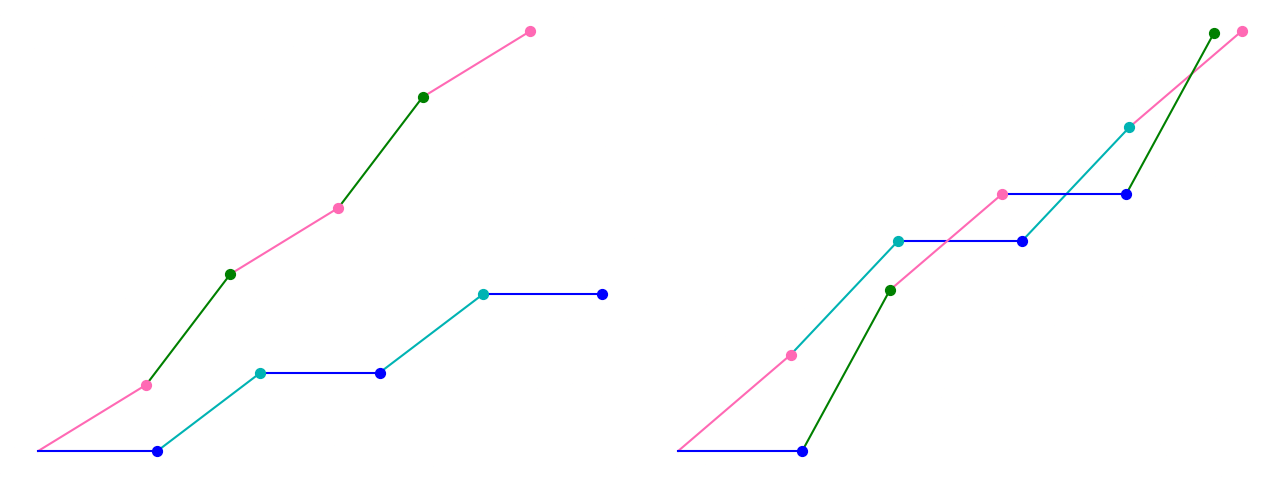}
        \put(-12, 38){ Parent Control Sequences} 
        \put(53, 38){ Children Control Sequences} 
        \put(75, 12){ \textcolor{myBlue}{$\delta$ = 0.000 rad}} 
        \put(75, 8){ \textcolor{myPink}{$\delta$ = 0.436 rad}}
        \put(75, 4){ \textcolor{myCyan}{$\delta$ = 0.524 rad}}
        \put(75, 0){ \textcolor{green}{$\delta$ = 0.785 rad}} 
    \end{overpic}
    \caption{Illustration of crossover. A visual showing how genetic crossover combines traits from two parent sequences to create new child sequences.}
    \label{fig:crossover}
\end{figure}
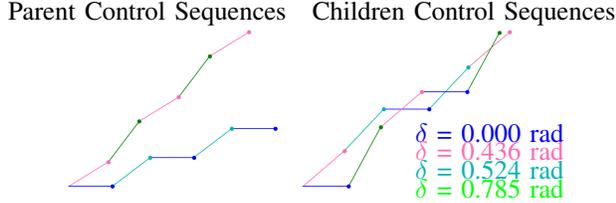

\item \textbf{Mutation:} Mutation is applied to each offspring based on a predefined mutation rate, introducing small changes to control sequences. This maintains genetic diversity and explores new regions of the search space. For each sequence \( u_i \) and control \( u_{i,t} \) at time \( t \), mutation adds a random perturbation \( \epsilon \) within the bounds of \( \mathcal{U} \), ensuring validity.
\begin{equation}
\begin{aligned}
\mathbf{u_{i,t}} & \leftarrow \mathbf{u_{i,t}} + \mathbf{\epsilon}, \\
& \quad \mathbf{u_{i,t}} + \mathbf{\epsilon} \in \mathcal{U}
\end{aligned}
\end{equation}
This heuristic approach guarantees that the changes introduced are valid.
    \item \textbf{Stopping Criteria:} The algorithm stops if the fitness of the best control sequence falls below a threshold (e.g., \( 10^{-3} \)) or if a predefined number of generations is reached. 
    
    \item \textbf{Elitism and Population Update:} At the end of each generation, the old population is replaced by the new population of offspring, which were produced through selection, crossover, and mutation.
    \item \textbf{Optimal Control Sequence:} Upon termination, the lowest-cost control sequence is chosen as the optimal policy for the finite horizon. The first step of this sequence is applied, and the process repeats with updated system states until the target is reached or the step limit is exceeded.
\end{itemize}

\begin{algorithm}
\caption{Genetic algorithm-based kinodynamic planning}
\label{alg:gaKDP}
\begin{algorithmic}
\State \textbf{\textcolor{myPink}{Input:}} Initial state $\mathbf{x_{init}}$, target state $\mathbf{x_{target}}$, timestep $\Delta t$, voxel grid $V$, mesh $M$, population size $N$, number of generations $G$, horizon $H$, mutation rate $\mu$
\State \textbf{\textcolor{myCyan}{Output:}} Optimal control sequence

\Function{\textcolor{myBlue}{GeneticAlgorithm}}{$\mathbf{x_{init}}$, $\mathbf{x_{target}}$, $\Delta t$, $V$, $M$, $N$, $G$, $H$, $\mu$}
    \State Generate initial population using \text{GeneratePopulation}
    \For{each generation $g \in [1, G]$}
        \State Compute fitness of each individual based on \ref{eqn:total_cost}
        \If{best fitness $< 10^{-8}$}
            \State \textbf{break}
        \EndIf
        \State Create new population by crossover and mutation
    \EndFor
    \State \Return the control sequence with the best fitness
\EndFunction

\Function{\textcolor{myBlue}{GAPlanner}}{start point, goal point, mesh}
    \State Initialize $\mathbf{x_{current}} \gets start point$
            \State trajectory $\gets [\mathbf{x_{current}}]$, and $\Delta t \gets 0.1$
    \For{each step up to \text{max\_steps}}
        \State Call \texttt{GeneticAlgorithm} $\to$ optimal control
        \State Update $\mathbf{x_{current}}$ based on \ref{eqn:dynamics}
        \State Append $\mathbf{x_{current}}$ to trajectory
        \If{$\| \mathbf{x_{current}} - \mathbf{x_{target}} \| < 0.1$}
            \State \Return trajectory
        \EndIf
    \EndFor
    \State \Return \texttt{None}
\EndFunction
\end{algorithmic}
\end{algorithm}
In summary, the methodology in Figure \ref{fig:abstract} and Algorithm \ref{alg:gaKDP} starts with an initial and target state. For each time step, we utilized a genetic algorithm to optimize the sequence of controls over a specified time horizon, assessing quality based on the cost function. 
Once the optimal path sequence was identified, we integrated the first step into the trajectory and repeated the process until the approximation to the target state was sufficiently close. The algorithm ultimately provides a set of path points that minimizes the cost of point-to-point transitions along the route from the initial to the target position.

\section{Evaluation}
The evaluation involved three key stages: first, we conducted a simulated comparison, assessing GAKD’s performance in a simulated environment and benchmarking it against two baseline algorithms, Model Predictive Path Integral (MPPI) and log-MPPI \cite{9834098}. 
Next, we implemented the planner using C++ in the ROS Move Base Flex framework, testing it with the botanical garden dataset from \cite{L36} and the Pluto Robot in a simulated 3D outdoor environment designed to replicate real-world conditions.
Finally, real-world testing was performed by deploying GAKD on a robot at Innopolis University, where we tested the planner on physical terrain to validate its effectiveness under practical conditions. 

The sections below  overview the evaluation metrics, experimental setup, and results of the comparisons, showcasing GAKD’s strengths across various scenarios.

\subsection{Evaluation Metrics}
The following metrics were used to evaluate planner performance:
\begin{itemize}
    \item \textbf{Traversability Metric:} Quantifies the difficulty of terrain navigation for a generated path. The cost of transitioning between states is estimated using \( \Pi \), as follows:
    $\Pi_{\text{metric}} = \frac{1}{s} \sum_{t=0}^s \Pi(\mathbf{x}_t, \mathbf{x}_{t+1}),$
    where higher values of \( \Pi_{\text{metric}} \) indicate poorer terrain traversability.
    \item \textbf{Path Length Metric:} Measures the deviation of the generated path length from the straight-line Euclidean distance between start and goal positions:
    $\Delta_{\text{metric}} = \left( \sum_{t=0}^{s} \Delta(\mathbf{x}_t, \mathbf{x}_{t+1}) \right) - \left\lVert \mathbf{x}_s - \mathbf{x}_0 \right\rVert$.
    \item \textbf{Trade-off Metric:} This is an affine combination of the min-max normalized \(\Delta_{\text{metric}}\) and $\Pi_{\text{metric}}$ metrics:
    $\Psi_{\text{metric}} = w\, \Pi_{\text{norm}} + (1-w)\, \Delta_{\text{norm}}$,
    where \(\Pi_{\text{norm}}\) and \(\Delta_{\text{norm}}\) are the normalized path efficiency and terrain traversability metrics across the scenarios, respectively, and \(w \in [0,1]\) adjusts their relative importance. Lower \(\Psi_{\text{metric}}\) indicates a more favorable trade-off.
\end{itemize}

\subsection{Experimental Design}
Two sets of experiments were conducted to evaluate the algorithms: 1). \textbf{Time Horizon Analysis:} Five start and goal configurations were tested, with the time horizon varied across values of 5, 7, 10, 12, and 15. This experiment assessed the impact of time horizon on path traversability and algorithm performance. 2). \textbf{Robustness and Consistency:} Using the same five start and goal configurations, a fixed time horizon of 10 was employed. Each configuration was tested over five trials to evaluate performance stability and robustness across repeated runs. Together, these experiments provided insights into the effects of time horizon on algorithm performance and its consistency across repeated trials.
\subsection{Experimental Setup}
\begin{figure}[htbp]
    \vspace{0.5cm}
    \centering
    \begin{overpic}[width=0.7\linewidth]{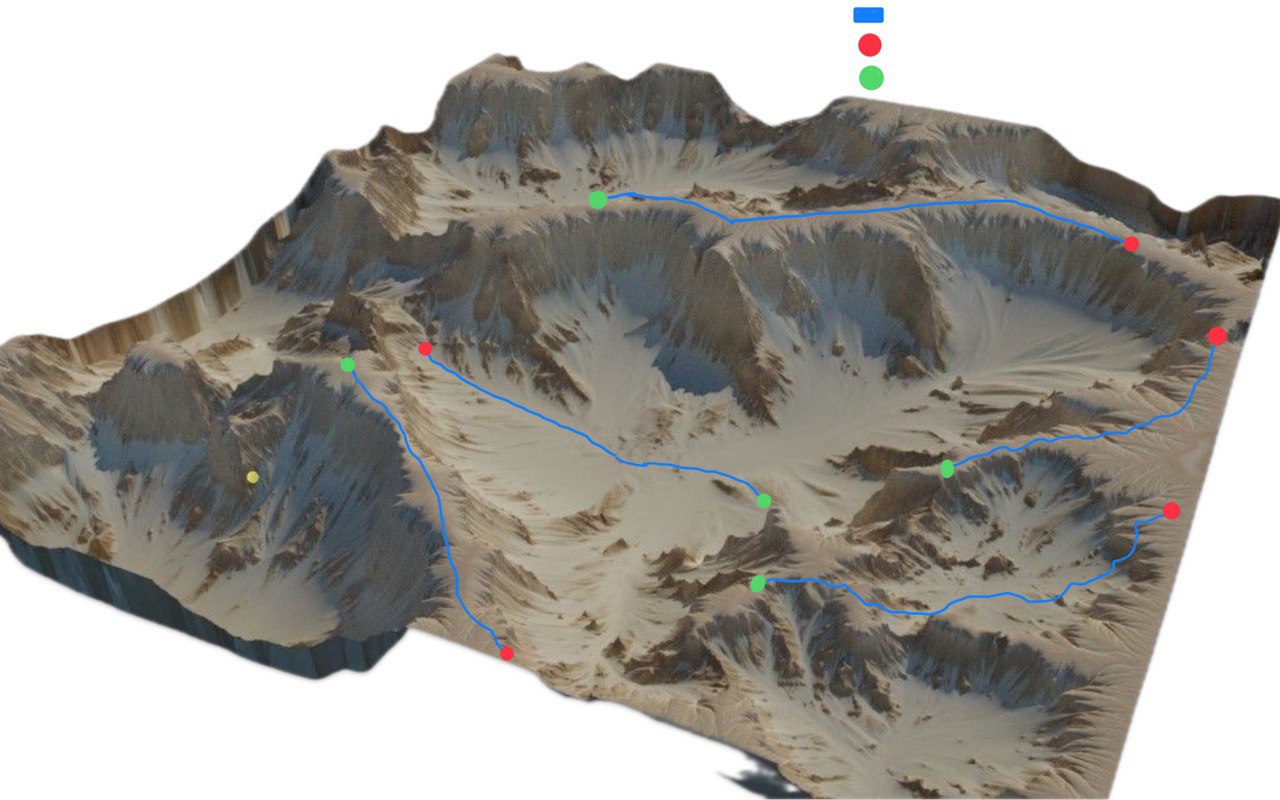}
        \put(70, 61){\scriptsize Path} 
        \put(70, 58){\scriptsize Start} 
        \put(70, 55){\scriptsize Stop}  
    \end{overpic}
    \caption{Textured view of the uneven terrain mesh and some test scenarios.}
    \label{fig:terrain}
\end{figure}

The experiments were conducted using Python implementations of GAKD, MPPI, and log-MPPI. Simulations were executed on an ASUS TUF A17 laptop equipped with an AMD Ryzen 7 4800H processor (8 cores, 16 threads, 2.90 GHz base frequency) and 8 GB of RAM. The PyVista library was used for 3D terrain visualization. The terrain in Figure \ref{fig:terrain} below is, obtained from \cite{cgtrader}, and contains a mesh with 1,552,238 vertices and 507,684 faces, offering a challenging environment for path planning. Identical system dynamics, acceleration limits, and steering angle constraints were applied across all algorithms to ensure fair comparisons.

\begin{table}[htbp]
\centering
\caption{Evaluation Scenarios}
\label{table:evaluation_scenarios}
\begin{tabular}{l c c c}
\hline
\textbf{Scenario} & \textbf{Start Point (x, y, z)} & \textbf{End Point (x, y, z)} & \textbf{Distance} \\
\hline
1 & -8.65, 6.13, -17.56 & 7.11, 3.64, 8.62 & 30.66 \\
2 & -7.52, 4.30, -15.82 & 5.52, 2.34, 5.98 & 25.48 \\
3 & -7.39, 4.42, -16.22 & 2.10, 1.88, -4.53 & 15.27 \\
4 & -8.98, 6.13, -17.19 & -1.23, 6.12, 10.06 & 28.34 \\
5 & -8.84, 6.11, -17.20 & -1.67, 6.17, 8.74 & 26.91 \\
\hline
\end{tabular}
\end{table}

\begin{figure}[htbp]
    \centering
    \includegraphics[width=0.7\linewidth]{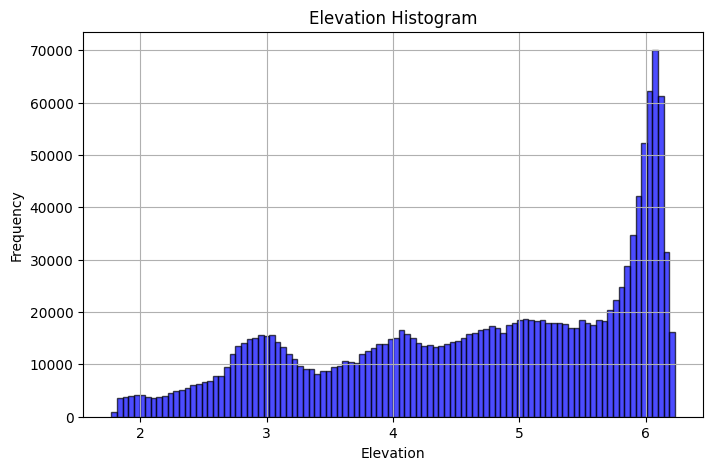}
    \caption{An elevation histogram of the mesh, shows the uneven distribution of height values across the surface.}
    \label{fig:elevation_hist}
\end{figure}

\subsection{Results and Analysis}

\subsubsection{Effect of Time Horizon}
\begin{figure}[htbp]
    \centering
    \includegraphics[width=0.8\linewidth]{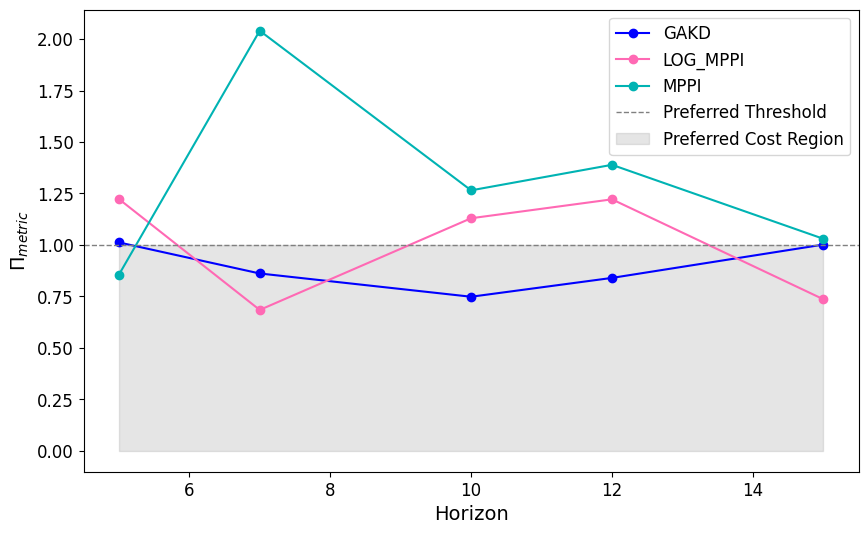}
    \caption{Average traversability cost by horizon. GAKD shows consistent performance across horizons, while MPPI and log-MPPI are more sensitive to horizon selection.}
    \label{fig:horizon}
\end{figure}

 GAKD demonstrated the least variability, with the best results between horizons of 7 and 12, as shown in Figure \ref{fig:horizon}. In contrast, MPPI performed best at a horizon of 5 but showed significant cost increases at longer horizons. log-MPPI exhibited high variability, achieving its lowest cost at horizon 7. Both MPPI and log-MPPI average control sequences and may accumulate errors, GAKD's crossover strategy creates a new control sequence by piecing together segments from different parent control sequences. This approach allows GAKD to completely replace segments with high cost, avoiding the error accumulation inherent in averaging.

\subsubsection{Traversability Across Various Scenarios}
\begin{table}[htbp]
\centering
\caption{Average traversability (\(\Pi_{\text{metric}}\)) for each planner across scenarios}
\label{table:pi_metric}
\begin{tabular}{l c c p{1.5cm}}
\hline
\textbf{Scenario} & \textbf{GAKD (\(\Pi_{\text{metric}}\))} & \textbf{log-MPPI (\(\Pi_{\text{metric}}\))} & \textbf{MPPI (\(\Pi_{\text{metric}}\))} \\
\hline
Scenario 1 & \cellcolor{cyan} 3.2900 & 5.2385 & 5.1999 \\
Scenario 2 & \cellcolor{cyan} 1.5352 & 3.0628 & 3.9137 \\
Scenario 3 & \cellcolor{cyan} 0.7483 & 1.1297 & 1.2653 \\
Scenario 4 & \cellcolor{cyan} 8.0621 & 8.4546 & 9.4023 \\
Scenario 5 & 9.5600 & \cellcolor{myPink}8.3866 & 9.3313 \\
\hline
\textbf{Average} & \textbf{4.6391} & \textbf{5.2544} & \textbf{5.8225} \\
\hline
\end{tabular}
\end{table}

\begin{figure}[htbp]
    \centering
    \includegraphics[width=0.8\linewidth]{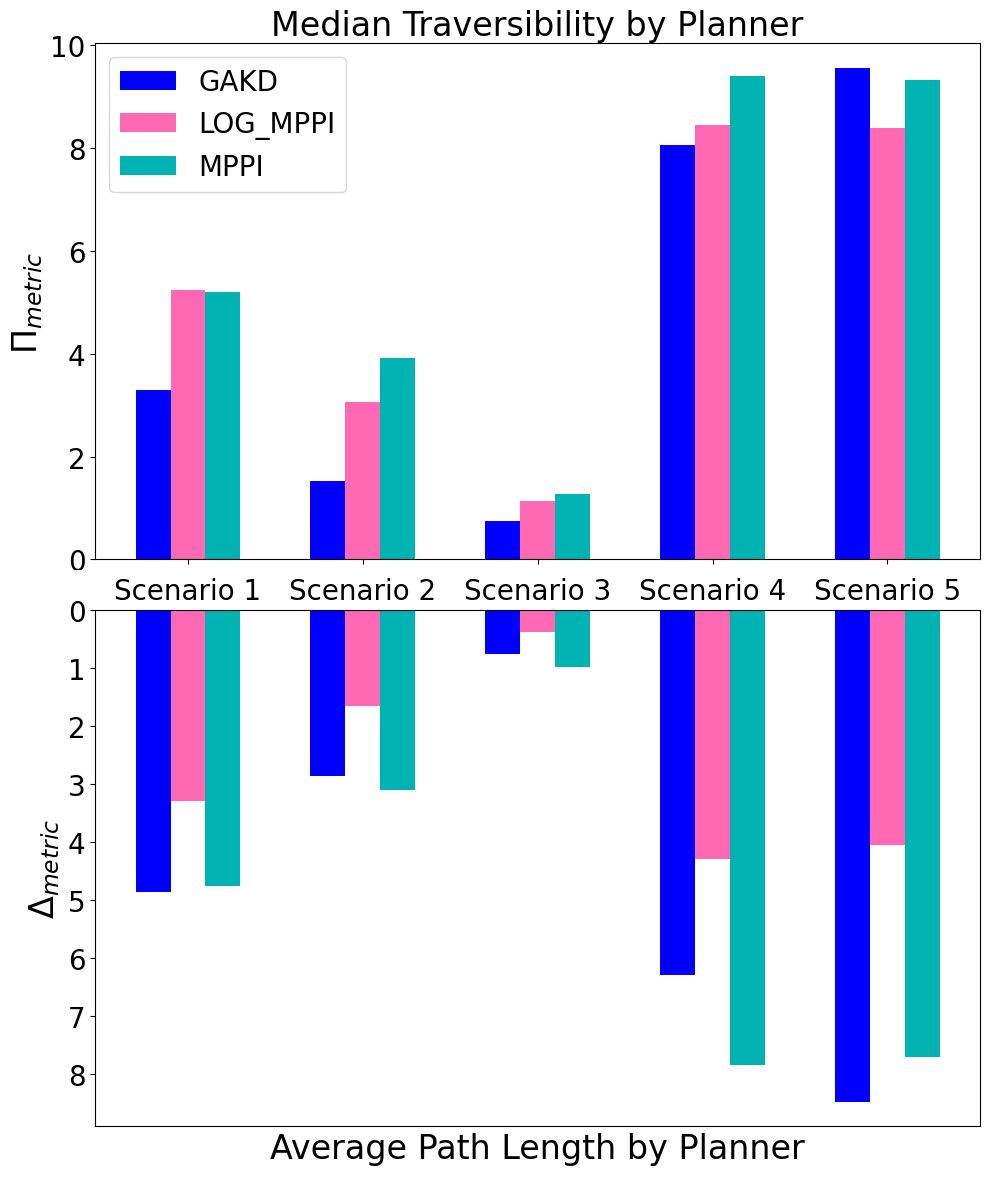}
    \caption{Comparison of algorithms based on path length and traversability cost across various scenarios. GAKD consistently outperforms both MPPI and log-MPPI in traversability cost, while log-MPPI achieves the shortest paths. GAKD demonstrates an effective balance between minimizing path length and maintaining low traversability costs.}
    \label{fig:scenarios}
\end{figure}

 GAKD achieved the lowest costs in most cases, as shown in Figure \ref{fig:scenarios} and Table \ref{table:pi_metric}, demonstrating its robustness across varying terrain difficulties. log-MPPI outperformed MPPI in lower-cost scenarios, but its variability made it less reliable overall.
 
\begin{table}[htbp]
\centering
\caption{Average path length (\(\Delta_{\text{metric}}\)) for each planner across scenarios}
\label{table:path_length}
\begin{tabular}{l c c c}
\hline
\textbf{Scenario} & \textbf{GAKD (\(\Delta_{\text{metric}}\))} & \textbf{log-MPPI (\(\Delta_{\text{metric}}\))} & \textbf{MPPI (\(\Delta_{\text{metric}}\))} \\
\hline
Scenario 1 & 4.8545 & \cellcolor{myPink} 3.2958 & 4.7641 \\
Scenario 2 & 2.8605 & \cellcolor{myPink} 1.6543 & 3.1074 \\
Scenario 3 & 0.7485 & \cellcolor{myPink} 0.3695 & 0.9742 \\
Scenario 4 & 6.2985 & \cellcolor{myPink} 4.2955 & 7.8432 \\
Scenario 5 & 8.4827 & \cellcolor{myPink} 4.0535 & 7.7075 \\
\hline
\textbf{Average} & \textbf{4.6489} & \textbf{2.7337} & \textbf{4.8793} \\
\hline
\end{tabular}
\end{table}

\begin{table}[ht]
\centering
\caption{(\(\Psi_{\text{metric}}\)) comparison of the planners (\(w =0.5\))}
\label{table:trade_off}
\begin{tabular}{l c c c}
\hline
\textbf{Scenario} & \textbf{GAKD (\(\Delta_{\text{metric}}\))} & \textbf{log-MPPI (\(\Delta_{\text{metric}}\))} & \textbf{MPPI (\(\Delta_{\text{metric}}\))} \\
\hline
Scenario 1 & 0.4097 & 0.6532 & 0.5176 \\
Scenario 2 & 0.1812 & 0.2956 & 0.3180 \\
Scenario 3 & 0.0 & 0.0 & 0.0 \\
Scenario 4 & 0.7738 & 1.0 & 1.0 \\
Scenario 5 & 1.0 & 0.9645 & 0.9858 \\
\hline
\textbf{Average} & \textbf{0.6729} & \textbf{0.5826} & \textbf{0.5643} \\
\hline
\end{tabular}
\end{table}

\subsubsection{Path Length Comparison}
GAKD generated paths of comparable length to MPPI, while log-MPPI often produced the shortest paths, as shown in Figure \ref{fig:scenarios} and Table \ref{table:path_length}. However, the trade-off between path length and traversability cost suggests that shorter paths do not always reduce traversability costs because focusing solely on path length can neglect challenging terrain, increasing overall traversal difficulty.  

It is shown that GAKD consistently outperformed the baselines across multiple metrics and scenarios. The average traversability cost (\(\Pi_{\text{metric}}\)) for GAKD was 4.6391, lower than both log-MPPI (5.2544) and MPPI (5.8225), demonstrating better terrain traversability. GAKD also exhibited the least variability in traversability across different time horizons. In contrast, MPPI showed significant cost increases for longer horizons, while log-MPPI displayed high variability. In terms of path length, GAKD’s average path length (\(\Delta_{\text{metric}}\)) was 4.6489, closely comparable to MPPI (4.8793), while log-MPPI produced the shortest paths on average (2.7337). However, GAKD’s ability to balance path length with traversability cost highlights its effectiveness in challenging terrains. These results demonstrate GAKD as an efficient alternative, especially in complex environments where terrain difficulty is critical.

\section{Real-World Demonstration}
To validate the proposed planner in a real-world environment, we deployed an Agilex Scout Mini robot equipped with a Velodyne VLP-16 LiDAR in an outdoor terrain at Innopolis University. The robot used POINT-LIO \cite{he2023point} for LiDAR-inertial odometry, generating a dense point cloud map. This map was subsequently converted into a triangular mesh in \texttt{.ply} format using the Las Vegas Reconstruction 2.0 (LVR2) toolkit~\cite{lvr2}. Additionally, terrain-specific layers were generated in \texttt{.h5} format using LVR2. These layers included surface roughness and elevation gradients.   

For navigation, we integrated the processed map into the MBF framework that computed static obstacles and the robot’s state representation. The proposed planner then used these inputs to compute a dynamically feasible trajectory. The trajectory was executed via MBF’s control interface, which sent velocity commands to the robot's onboard controller. The MBF setup used is available at:  \url{https://github.com/Stblacq/mesh_planner}.

To evaluate performance, we conducted four test runs across five scenarios, measuring computation time, traversability, and path length. The computation time \(\mathcal{T}\) represents the average time required to compute a trajectory. The results are summarized in Table~\ref{tab:planner_comparison} showing that MPPI has the highest mean computation time, followed closely by GAKD, with relatively low variability. In contrast, log-MPPI demonstrates the fastest mean computation time but higher variability.

\begin{table}[htbp]
    \centering
    \caption{Average performance metrics for each planner}
    \label{tab:planner_comparison}
    \begin{tabular}{l c c c}
        \hline
        \textbf{Metric} & \textbf{GAKD} & \textbf{log-MPPI} & \textbf{MPPI} \\
        \hline
        \(\mathcal{T}\) [s] & 7.60 $\pm$ 0.10 & 7.35 $\pm$ 0.38 & 7.65 $\pm$ 0.12 \\
        \(\Pi_{\text{metric}}\) & 4.6391 $\pm$ 0.08 & 5.2544 $\pm$ 0.30 & 5.8225 $\pm$ 0.10 \\
        \(\Delta_{\text{metric}}\) & 4.6489 $\pm$ 0.08 & 3.7337 $\pm$ 0.28 & 4.8793 $\pm$ 0.10 \\
        \hline
    \end{tabular}
\end{table}

\section{Discussion}
One caveat of the proposed GAKD algorithm is that this algorithm is heuristic-based and relies on user expertise for selecting parameters such as \(\alpha_1, \alpha_2\), as well as for designing the dynamic model and cost function. However, this also provides flexibility because different dynamic models and cost functions can be used depending on the application. 
Theoretically, the robot state and mesh changes within the planning horizon can be used as input in a dynamic environment. The cost function can be adapted to minimize undesired features for robust evaluation on uneven surfaces.

\section{Conclusion and Future Work}
This study presents GAKD, designed to optimize trajectory planning for car-like vehicles navigating uneven terrains modeled with triangular meshes. Using heuristic-based mutation within a genetic algorithm, the planner effectively maintains control within the vehicle's operational range.
The experimental evaluations demonstrated that GAKD consistently outperformed the baseline algorithms, MPPI and log-MPPI, across critical metrics and scenarios. Specifically, GAKD achieved the lowest average traversability cost (\(\Pi_{\text{metric}} = 4.6391\)), indicating its improved ability to navigate challenging terrains compared to MPPI (5.8225) and log-MPPI (5.2544). Furthermore, GAKD exhibited minimal variability in performance across different time horizons.
In terms of path length, GAKD achieved an average path length (\(\Delta_{\text{metric}} = 4.6489\)) comparable to MPPI (4.8793) while maintaining lower traversability costs, which highlights its effectiveness in balancing path length and terrain adaptability. Although log-MPPI generated the shortest paths (\(\Delta_{\text{metric}} = 2.7337\)), its higher variability and focus on path length often neglected traversability, leading to higher costs under challenging terrains.
These results validate GAKD as an effective kinodynamic planning solution, particularly in scenarios requiring navigation through complex and uneven terrains.
Future work will explore simulated annealing and Simultaneous Perturbation Stochastic Approximation (SPSA) for enhancing motion planning, as well as data-driven approaches to generating dynamic models.

\bibliographystyle{IEEEtran}
\bibliography{IEEEabrv,ref}
\end{document}